\documentclass[letterpaper,10pt,conference]{ieeeconf}
\IEEEoverridecommandlockouts
\usepackage[T1]{fontenc}
\usepackage{mathptmx}
\usepackage{amsmath,amssymb}
\usepackage{booktabs,tabularx,array}
\usepackage{multirow}
\usepackage{cite}
\usepackage{url}
\usepackage{graphicx}
\usepackage{xcolor}
\usepackage[nolist,nohyperlinks]{acronym}
\usepackage{tikz}
\usetikzlibrary{arrows.meta,positioning,calc}
\newcolumntype{Y}{>{\raggedright\arraybackslash}X}
\newcommand{\R}{\mathbb{R}}
\newcommand{\pos}[1]{\left[#1\right]_+}

\newcommand{\ikpd}{\mbox{IK\,+\,PD}}

\acrodef{IPT}{Interaction Primitive Training}
\acrodef{RL}{reinforcement learning}
\acrodef{QDD}{Quasi-Direct-Drive}
\acrodef{IK}{inverse kinematics}
\acrodef{PD}{proportional--derivative}
\acrodef{Hybrid}{hybrid motion--force control}
\acrodef{Impedance}{Cartesian impedance control}
\acrodef{NRIC}{nonlinear robust internal-loop compensator}
\acrodef{PPO}{proximal policy optimization}
\acrodef{MLP}{multilayer perceptron}
\acrodef{ELU}{exponential linear unit}
\acrodef{VLA}{vision--language--action}
\acrodef{IL}{imitation learning}
\acrodef{MIT}{Massachusetts Institute of Technology}
\acrodef{FORGE}{Force-Guided Exploration}
\acrodef{EE}{end-effector}
\title{\LARGE\bfseries Learning Reflexive Behavior for Contact-Rich Manipulation}
\author{Quan Nguyen, Yunho Kim, Joonho Lee$^{*}$\\
{\normalsize All authors are with Neuromeka Co., Ltd.}\\
{\normalsize * Corresponding author. Email: \texttt{joonho.lee@neuromeka.com}.}
}
\date{}

\begin{document}
\maketitle
\thispagestyle{empty}
\pagestyle{empty}

\begin{abstract}
During contact-rich manipulation, interactions between a robot and its environment carry information about local geometry: a surface prevents penetration, a bore guides a peg. A controller that exploits  these interactions can can comply with environmental constraints while preserving task intent. 
Robot-learning systems commonly use position, hybrid force–position, or Cartesian impedance control. Their prescribed tracking objectives, stiffness, or force-control directions may not match local constraints and may degrade performance.
We learn a proprioceptive reflex policy in simulation on three simple interaction primitives: a spring, a plane, and a rail.
The policy maps task-space commands to joint-position targets using state history, without direct force or geometrical measurements.
Once trained, it serves as a frozen execution layer beneath higher-level controllers.
We evaluate it in dual-arm box lifting, peg insertion, and surface following. 
In box lifting, the reflex kept the force below the threshold while the baseline failed. In rough-surface following it stayed below the 10 N reference, on par with tuned hybrid force-position control. In 0.02 mm peg insertion It reduced mean estimated contact force to less than half that of the baseline while increasing hardware success rates from at most 22 \% to 36–58 \%. By separating contact response from command generation, the reflex policy provides motion planners, learned policies, and teleoperators with robust contact-rich execution.
\end{abstract}

\acresetall


\section{Introduction}

In contact-rich manipulation, environmental geometry shapes how a robot or its tools can move and provides guidance for completing a task. A surface supports sliding while restricting penetration, and a groove guides motion along a path. By maintaining contact while yielding in constrained directions, a robot can exploit this geometry to guide execution despite errors in perception, grasping, or nominal motion.

\begin{figure}[!t]
\centering
\includegraphics[width=\columnwidth]{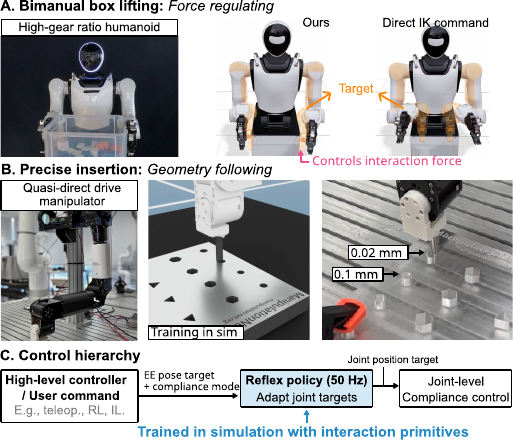}
\par\medskip
\input{figures/interaction_priors}
\caption{Overview. 
(A)~Bimanual box lifting on the humanoid (left), and joint-position targets (orange) against the measured robot state (solid) with the reflex (center) and with the nominal \ac{IK} targets (right). The reflex pulls the targets back toward the box surface, which limits the squeezing force, whereas the nominal targets lie deep inside the box.
(B)~Peg insertion: the \ac{QDD} arm (left), training in simulation (center), and the evaluation holes with 0.02 and 0.1\,mm tolerances (right).
(C)~Control hierarchy with the reflex between the task-level command source and the joint-level controller. Bottom: the three interaction primitives used in training: (a)~spring, (b)~plane, and (c)~rail.}
\label{fig:opening}
\label{fig:interaction_priors}
\vspace{-5mm}
\end{figure}

High-level commands from a motion planner, a learned policy, or a human operator may specify the intended behavior without matching the exact geometry of the current task instance~\cite{li2026single, shi2026minimalist, kim2026learning}.
For example, a surface-search command is useful only if execution maintains contact while moving over the actual surface. Li and Brock~\cite{li2026single} separate task structure from instance-specific geometry and recover the latter through interaction with environmental constraints. Compliant execution beneath high-level policies similarly supports adaptation to local contact~\cite{shi2026minimalist}. These observations motivate an execution capability that makes environmental constraints useful to a broad range of command sources.

We present a method for training a proprioceptive reflex policy through \ac{RL} using three primitive interaction scenarios shown in Fig.~\ref{fig:opening} (bottom).
The proposed a control hierarchy is shown in Fig.~\ref{fig:opening}C with high level controller supplies contact intent and nominal motion while the reflex policy adjusts joint-space targets using a history proprioceptive measurements, and a compliant joint controller executes them. A single training environment samples virtual spring, plane, and rail constraints alongside free-space motion. The objective encourages progress toward a compliant reference and command-derived interaction force while penalizing excessive loading. This formulation builds directly on three prior designs: the force-threshold penalty of \ac{FORGE}~\cite{noseworthy2025forge}, through which an \ac{RL} policy learns to respect a maximum allowable force; the force-compensated position reward of Zhi et al.~\cite{zhi2025unified}; and the interaction-mode command of Lee et al.~\cite{lee2025toolaware}, which tells a controller whether contact is permitted.




We evaluate our approach across the two robot embodiments shown in Fig.~\ref{fig:opening}. We apply the learned reflex controller to improve performance on dual-arm box lifting (A) and high-precision peg insertion (B) tasks. The results show that the learned reflex supports stable lifting under imperfect motion commands, enables insertion under hole-position uncertainty, and follows rough surfaces while limiting contact loading. Our contribution is a reusable reflex-training paradigm that uses simple interaction primitives to train policies that adapt task-level motion commands to local contact.


\section{Related Work}
\subsection{Environmental Constraints and Compliant Manipulation}
Li and Brock~\cite{li2026single} represent manipulation with environmental constraint (EC) primitives, recover instance geometry through exploratory interaction, and execute through compliant contact; their plane primitive, for example, maintains surface contact during motion and search. Minimalist Compliance Control~\cite{shi2026minimalist} places actuator-based wrench estimation and task-space admittance beneath visual, imitation, and model-based policies. Both separate task intent from the response to local contact, the division our reflex adopts. Impedance control specifies a mechanical relationship between motion error and interaction force~\cite{hogan1985impedance}, and admittance control turns measured force into reference motion for an inner position loop~\cite{ott2010unified}; these classical schemes are the reference points against which we position a learned reflex. We ask whether learning from structured interactions improves on them when the nominal geometry is imperfect.

\subsection{Learned Interaction Control}
\ac{FORGE}~\cite{noseworthy2025forge} conditions assembly policies on a maximum allowable contact force and penalizes simulated force above that threshold during \ac{RL}.The policy learns to modulate its actions so that the threshold is respected regardless of controller gains.
This force-thresholded \ac{RL} formulation is the central insight our training objective shares: the excess-force cost in Sec.~\ref{sec:simulation} plays the role of the \ac{FORGE} penalty, and the peg-insertion study (Sec.~\ref{sec:downstream}) builds on its assembly setup. Portela et al.~\cite{portela2024force} and Zhi et al.~\cite{zhi2025unified} learn force or position--force control for legged manipulators from robot-state history without force sensors. Zhi et al. reward tracking of a target position that is offset from the command by the net force through a static impedance relation, so that position tracking, force application, and compliance follow from one reward. Our compliant reference reward is based on this reward design. Direction Matters~\cite{yang2026direction} predicts a contact-force direction for admittance control. Lee et al.~\cite{lee2025toolaware} supply a collision-avoidance controller with a binary interaction-mode command that determines whether tool contact is permitted. Inspired by this design, we use a compliance-mode input to tell the policy whether to prioritize following the commanded motion or yielding to contact forces. Unlike these systems, the reflex policy adapts joint-space references beneath an independently supplied task command.

\subsection{Proprioceptive Control and Local Reflexes}
Proprioceptive control and local reflexes have been studied as mechanisms for adapting robot motion during physical interaction. Hyun et al.~\cite{hyun2014proprioceptive} combine high-level gait commands with proprioceptive impedance control on the MIT Cheetah, while Wensing et al.~\cite{wensing2017proprioceptive} connect effective physical interaction to backdrivable actuation and low mechanical impedance. In manipulation, SaLoutos et al.~\cite{saloutos2022reflexive} use local anti-slip and regrasping reflexes to correct imperfect teleoperator grasps. Their reflexes use fingertip force/contact sensing; our reflex policy observes joint and kinematic history. These results motivate learning proprioceptive reflexes from environmental interaction and evaluating their reuse across manipulation conditions.

\section{Method}

\label{sec:method}
As in operational-space control, where a contact task is characterized by its constrained and free directions~\cite{khatib1987operational}, we approximate environmental constraints with three interaction primitives—spring, plane, and rail—that apply virtual forces to the simulated robot's \ac{EE} (Sec.~\ref{sec:priors}). 
Through \ac{RL}, we train a proprioceptive reflex policy to follow task-space commands while accommodating these interactions (Sec.~\ref{sec:simulation}). The policy uses command and proprioceptive history to generate joint-position targets. 
During training, the sampled interaction primitives define compliant position and interaction-force targets, while a separate penalty discourages excessive loading.

\subsection{Interaction Primitives}
\label{sec:priors}

Each interaction primitive is a virtual spring--damper model that computes an external force from the \ac{EE}'s displacement and velocity and applies it to the simulated robot; the bottom row of Fig.~\ref{fig:opening} shows the three primitives in simulation. The models below define the forces, and the reflex policy learns to keep them below the threshold while maintaining the commanded motion.

\paragraph{Spring primitive}
The spring pulls the \ac{EE} toward an anchor position $x_a$ in the workspace with the force
\begin{equation}
 F^{\mathrm s}=k_{\mathrm s}(x_a-x)-b_{\mathrm s}\dot{x},
 \label{eq:spring}
\end{equation}
where $x$ is the \ac{EE} position and $k_{\mathrm s}$ and $b_{\mathrm s}$ are the spring stiffness and damping coefficient. We sample the anchor position uniformly within the workspace.

\paragraph{Plane primitive}
In tasks such as surface wiping or deburring, the robot moves along a surface while pressing against it. We model this with a virtual plane that pushes back whenever the \ac{EE} penetrates it. Let $d(x)$ be the signed distance from the \ac{EE} to the plane where $d(x) < 0$ means penetration, and $v_n$ be the normal velocity. The force magnitude is described as
\begin{equation}
 f^{\mathrm p}=\begin{cases}
 0, & d(x)\geq 0,\\
 -k_{\mathrm p}d(x)-b_{\mathrm p}v_n, & d(x)<0,
 \end{cases}
 \label{eq:plane}
\end{equation}
where $k_{\mathrm p}$ and $b_{\mathrm p}$ are the plane stiffness and damping coefficient. The force acts only along the plane normal, leaving motion across the surface free.

\paragraph{Rail primitive}
In tasks such as peg insertion, the surrounding surfaces constrain two translational directions and leave motion along the remaining direction free. We model this with a virtual rail that pulls the \ac{EE} back toward a fixed straight line. Let $d_g$ be the shortest distance from the \ac{EE} to the rail and $v_g$ be its rate of change. The force magnitude is described as
\begin{equation}
 f^{\mathrm g}=k_{\mathrm g}d_g+b_{\mathrm g}v_g,
 \label{eq:guide}
\end{equation}
where $k_{\mathrm g}$ and $b_{\mathrm g}$ are the rail stiffness and damping coefficient. The force direction points from the \ac{EE} toward the rail is always perpendicular to the rail.


\subsection{Reflex Policy Training}
\label{sec:simulation}
We train the reflex policy in simulation using the primitives above. The policy tracks a commanded \ac{EE} pose from a short proprioceptive history (0.5 s) and outputs joint-position targets at 50 Hz. Here we define the episode setup, the observation and action spaces, and the reward function.

Each episode starts from a reachable \ac{EE} pose drawn from precomputed \ac{IK} solutions and uses either one primitive or free-space motion. Similarly to~\cite{lee2025toolaware}, a binary compliance-mode input $m_t$ is one for primitive episodes and zero otherwise, indicating whether the policy should yield to contact or track the command exactly.

\subsubsection{Action}
The action $a_t\in\R^n$ is a residual joint-position target with fixed per-joint scales $\sigma_a$:
\begin{equation}
 q_t^{\mathrm{ref}}=q_t+\sigma_a\odot a_t,
 \label{eq:action}
\end{equation}
where $\odot$ denotes the element-wise product. The low-level controller tracks $q_t^{\mathrm{ref}}$. Gaussian action distribution is used for stochastic \ac{RL}.

\subsubsection{Observation}

The observation is a 25-step (0.5\,s) history of joint positions $q_\tau\in\R^n$, joint-target errors $e_\tau^q=q_\tau^{\mathrm{ref}}-q_\tau$, commanded and measured \ac{EE} poses $p_\tau^d$ and $p_\tau$, and the compliance mode $m_\tau$, i.e., $o_t=[q_\tau,e_\tau^q,p_\tau^d,p_\tau,m_\tau]_{\tau=t-24}^{t}$. Here $n$ is the number of joints (7 per arm); with two arms, per-arm quantities are concatenated. Poses are expressed in the base frame as a position and the continuous 6D rotation representation~\cite{zhou2019continuity}. The policy receives neither force measurements nor primitive parameters.


\begin{table}[t]
\caption{Reward terms for reflex policy training. The reward is the weighted sum of all terms; negative weights denote penalties. Position errors are measured to the compliant reference $x_t^c$ (Eq.~\eqref{eq:reference_projection}). Force terms are in newtons and apply only in compliance mode ($m_t=1$).}
\label{tab:reward_terms}
\centering
\footnotesize
\setlength{\tabcolsep}{4pt}
\begin{tabularx}{\columnwidth}{@{}lXr@{}}
\toprule
Term & Formula & Weight \\
\midrule
\multicolumn{3}{@{}l}{\textit{Pose tracking}} \\
Position, coarse & $\kappa(\|x_t-x_t^c\|/0.1)$ & $0.10$ \\
Position, fine & $\kappa(\|x_t-x_t^c\|/0.025)$ & $0.05$ \\
Orientation & $\kappa(\theta_t/10^\circ)$ & $0.20$ \\
\addlinespace
\multicolumn{3}{@{}l}{\textit{Interaction force}} \\
Force tracking & $|f_t-f_t^d|$ & $-0.02$ \\
Excess force & $\pos{f_t-F_{\mathrm{th}}}$ & $-0.02$ \\
\addlinespace
\multicolumn{3}{@{}l}{\textit{Regularization}} \\
\ac{EE} speed & $\|\nu_t\|$ & $-10^{-3}$ \\
Joint speed & $\|\dot q_t\|^2$ & $-10^{-3}$ \\
Action, 1st diff. & $\|\Delta a_t\|^2$ & $-5\times10^{-5}$ \\
Action, 2nd diff. & $\|\Delta^2 a_t\|^2$ & $-5\times10^{-5}$ \\
Target change & $\|\sigma_a^{-1}\odot\Delta q_t^{\mathrm{ref}}\|^2$ & $-5\times10^{-5}$ \\
Joint limits & $c_{\mathcal Q}(q_t)$ & $-10^{-2}$ \\
\midrule
\multicolumn{3}{@{}p{\columnwidth}@{}}{%
$\kappa(u)=(1+u^2)^{-1/2}$; 
$\ell_{\mathrm c}>\ell_{\mathrm f}$: coarse and fine length scales; 
$\theta_t$: angle between measured and commanded orientations; 
$f_t$: magnitude of the interaction primitive's force; 
$\nu_t$: \ac{EE} velocity; 
$\Delta$: backward difference, $\Delta a_t=a_t-a_{t-1}$; 
$c_{\mathcal Q}$: joint-limit violation cost.} \\
\bottomrule
\end{tabularx}
\end{table}

\subsubsection{Reward}
The reward combines task-space pose tracking and force-related terms; Table~\ref{tab:reward_terms} lists their definitions and weights.
\begin{itemize}
 \item \textbf{Force-projected position tracking reward.} In compliance mode, the policy should track the commanded position $x_t^d$ only as far as the interaction force allows. Building on Zhi et al.~\cite{zhi2025unified}, we define a compliant reference $x_t^c$ as the projection of $x_t^d$ onto a force-feasible set. Let $k$ be the stiffness of the sampled interaction primitive and $\delta(x)$ its displacement at position $x$,
 \begin{equation}
  \delta(x)=\begin{cases}
  \|x-x_a\|, & \text{spring},\\
  \pos{-d(x)}, & \text{plane},\\
  \|x-x_\Gamma(x)\|, & \text{rail},
  \end{cases}
  \label{eq:compliant_reference}
 \end{equation}
 so that the static force is $k\delta(x)$. Here $x_\Gamma(x)$ is the nearest point on the rail and $[u]_+=\max(u,0)$. The force-feasible set and compliant reference are
 \begin{equation}
  \mathcal C=\{x:k\delta(x)\leq F_{\mathrm{th}}\},\qquad
  x_t^c\in\underset{x\in\mathcal C}{\arg\min}\,\|x-x_t^d\|^2.
  \label{eq:reference_projection}
 \end{equation}

$x_t^c$ and $x_t^d$ become equal when the static force at $x_t^d$ is at most $F_{\mathrm{th}}$. In another words, we modify the position target by translating it along the force direction until the calculated static force becomes smaller than $F_{th}$. This target modification only apply when the compliance mode is set to one. And the position-tracking rewards use this $x_t^c$ in place of $x_t^d$ as the target. We do not modify the orientation target.

 \item \textbf{Interaction force reward.} We have two interaction force reward terms. First we define the force reference as the force that the \ac{EE} would experience when it  commanded position, capped 1\,N below the threshold:
 \begin{equation}
  f_t^d=\min\!\left(k\,\delta(x_t^d),F_{\mathrm{ref}}\right),\qquad
  F_{\mathrm{ref}}=F_{\mathrm{th}}-1\,\mathrm{N},
  \label{eq:force_reference}
 \end{equation}
 where the margin reduces threshold violations caused by tracking errors. The force-tracking term penalizes the absolute error $|\bar f_t-f_t^d|$, where $\bar f_t$ is the mean magnitude of the interaction primitive's force over one policy interval. The excess-force term adds the violation $\pos{\bar f_t-F_{\mathrm{th}}}$, which steepens the penalty above the threshold. Both terms are disabled when $m_t=0$ or no interaction primitive is active.
\end{itemize}

\subsubsection{Training details}
We train the policy with \ac{PPO}~\cite{schulman2017ppo} in an asymmetric actor--critic setup~\cite{pinto2017asymmetric}. The actor encodes the 25-step observation history with two temporal convolution layers (32 channels, kernel size 3) into a 512-dimensional latent. An \ac{MLP} with hidden widths 512 and 256 and \ac{ELU} activations then parameterizes the Gaussian policy. The critic shares this structure and additionally receives the ground-truth interaction force as privileged information, which the deployed actor never observes. The simulation runs at 150\,Hz, i.e., three physics steps per policy step, with high position and velocity solver iteration counts.

\section{Results}
\label{sec:results}

We evaluate the reflex policy on three tasks: dual-arm box lifting, peg insertion, and surface following. Commands come from scripted trajectories or a learned high-level policy, and experiments run in simulation and on two robots with different actuation. We ask two questions: (Q1)~When commands are imperfect, such as excessive commanded penetration or an erroneous hole position, does the reflex policy limit contact force and sustain task success where conventional execution layers fail? (Q2)~Is this benefit simply a matter of compliance, or does it require the constraint structure of the interaction primitives? We address Q2 from two sides: classical compliant controllers swept over gains, and reflex variants trained with one primitive removed. Additional details and videos are provided in the supplementary material.

We compare against three classical execution layers. \ikpd{}, the baseline in every experiment, converts task-space commands to joint targets with \ac{IK} and tracks them with each robot's joint-level controller (Sec.~\ref{sec:robots}). Impedance is a Cartesian impedance controller, used in the insertion and surface following studies. Hybrid is hybrid motion--force control, used in the surface following study.

\subsection{Robots and low-level controllers}
\label{sec:robots}

We use two robots with different actuator types. Both track the policy's joint-position targets with a compliant joint-level controller, but the compliance has different origins: on the high-gear-ratio humanoid it is rendered by model-based control, and on the \ac{QDD} manipulator it comes from the backdrivability.

\begin{itemize}
 \item \textbf{Industrial humanoid with model-based compliance control} (Fig.~\ref{fig:opening}A). The box-lifting experiment uses a humanoid robot with two 7-DoF arms. Its sensorless joint-space compliance controller, based on Ko et al.~\cite{ko2023bounded}, gives the harmonic-drive joints \ac{PD}-like compliant behavior. To reduce the sim-to-real gap, we identify joint damping, friction, delay, and reflected inertia with PACE~\cite{bjelonic2025towards}.

 \item \textbf{Custom manipulator with \ac{QDD} actuators} (Fig.~\ref{fig:opening}B). 
 The peg-insertion experiments use a custom 7-DoF arm derived from OpenArm~\cite{openarm}: it keeps OpenArm's \ac{QDD} actuators but has modified kinematics. Its low gear ratios, inertia, and friction make it highly backdrivable. The joint-level controller is a \ac{PD} law with feedforward torque~\cite{mitcheetahsoftware},
$\tau^{\mathrm{cmd}}=K_p e+K_d\dot e+\tau^{\mathrm{ff}}$ with $e=q^{\mathrm{ref}}-q$, where $K_p$, $K_d$, and $\tau^{\mathrm{ff}}$ are the joint stiffness, damping, and feedforward torque.
\end{itemize}

\ikpd{} baseline solves \ac{IK} with the Pink solver~\cite{pink} and sends the resulting joint-position targets directly to the low-level controllers above.
Impedance and Hybrid use the controller implementations provided in Isaac Lab~\cite{mittal2025isaaclab} that follows~\cite{khatib1987operational}.


\subsection{Dual-arm Robot Box Lifting}
\label{sec:lifting}

\begin{figure}[t]
\centering
\includegraphics[width=\columnwidth]{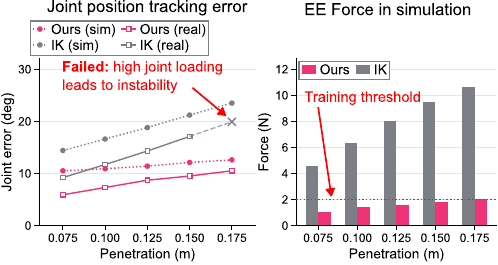}
\caption{Box lifting with the reflex and \ikpd{} at increasing commanded penetration depths. Left: average joint position tracking error in simulation and on hardware; a larger error means that the joint targets lie farther inside the box (orange in Fig.~\ref{fig:opening}A). \ikpd{} diverged on hardware at 17.5\,cm. Right: simulated palm contact force; the dotted line marks the 2\,N training threshold.}
\label{fig:box_lifting_exp}
\vspace{-5mm}
\end{figure}

In bimanual lifting, the two palms press on opposite faces of the box, and the squeezing force must stay within what the arms and the box can safely sustain. Under position control, this force is set indirectly: it scales with the controller stiffness and with how far the \ac{EE} targets lie inside the box, so an error in commanded grasp width becomes squeezing force. We use this task to address Q1 with a single controlled source of command error, the commanded penetration depth.

\subsubsection{Experimental setup}
A rigid 1\,kg box ($28\times36\times17$\,cm) is placed at the same pose in every trial (Fig.~\ref{fig:opening}A). (The box has a latch on each grasped face that rests on the hands during the lift, so the lift does not rely on friction alone.)

A scripted task-space trajectory moves both hands to the side faces, commands each hand inward by a penetration depth $d_{\mathrm{pen}}$ from its contact surface, and lifts the box by 10\,cm. We test $d_{\mathrm{pen}}\in\{7.5,10,12.5,15,17.5\}$\,cm with the reflex policy and with \ikpd{}. Both receive the same trajectory and run on the same joint-level controller ($K_p=20$\,N\,m/rad and $K_d=5$\,N\,m\,s/rad for all joints). The reflex policy was trained with $F_{\mathrm{th}}=2$\,N, and its compliance mode is held at $m_t=1$ throughout.

The robot has no force or joint-torque sensors, so we report two quantities: (i)~the average joint position tracking error on hardware, i.e., $\|q_t^{\mathrm{ref}}-q_t\|$ averaged over a 30\,s hold, and
(ii)~the palm contact force in simulation, obtained by replaying the same trajectories with the same robot and box models. Because the joint-level controller behaves like a \ac{PD} law, this error is roughly proportional to the steady-state joint torque and serves as a hardware proxy for grasp loading.

\subsubsection{Results}
The reflex policy lifted the box at all five depths. \ikpd{} lifted it at the first four but failed at 17.5\,cm: under the grasp load produced by this command, the low-level controller diverged and the arms vibrated strongly.

On hardware, the average joint position tracking error of \ikpd{} grew steadily with commanded depth (Fig.~\ref{fig:box_lifting_exp}, left), because \ac{IK} places the joint targets at the commanded pose inside the box regardless of contact (Fig.~\ref{fig:opening}A, right). With the reflex, the error stayed below $12^\circ$ over the full range: the policy holds the joint targets near the box surface instead of following the command into the box (Fig.~\ref{fig:opening}A, center), so additional commanded depth does not translate into joint torque.

In simulation, the palm contact force under \ikpd{} rose monotonically with depth, whereas the reflex kept it at or below the 2\,N training threshold at every depth (Fig.~\ref{fig:box_lifting_exp}, right). These forces were not measured on the robot, but the simulation reproduces the hardware trends in tracking error (Fig.~\ref{fig:box_lifting_exp}, left), which supports the trend in force, though not its absolute values.


\subsection{Contact-Guided Peg Insertion}
\label{sec:downstream}

Tight-clearance insertion cannot be solved by positioning alone: the clearance (0.02--0.1\,mm) is orders of magnitude smaller than the error in the estimated hole position, so the peg must find and enter the hole through contact. 
Under a stiff position control, the same position error turns into large lateral forces at the rim, where the peg can jam. 

In this study the commands come from a learned policy instead of a script: we train a high-level insertion policy on top of the frozen reflex and compare it with policies trained on top of classical controllers. The study addresses Q1, with hole-position error as the command error, and Q2, through the Impedance baseline and the primitive ablations.


\begin{table}[t]
\caption{Real-world peg insertion with the reflex (ours) or \ikpd{} as the execution layer beneath a LSTM insertion policy.}
\label{tab:insertion_real_world}
\centering\small
\setlength{\tabcolsep}{3pt}
\begin{tabular*}{\columnwidth}{@{\extracolsep{\fill}}lrrrrr@{}}
\toprule
& Offset & Success & Time (s) & \multicolumn{2}{c@{}}{Est. force (N)} \\
\cmidrule(l){5-6}
Method & (mm) & count (\%) & Mean $\pm$ SD & Mean & Peak \\
\midrule
\multicolumn{6}{@{}l}{\textit{0.02\,mm tolerance}} \\
\multirow{4}{*}{Reflex}
 & 0 & 29/50 (58) & $10.47\pm4.40$ & 7.41 & 34.01 \\
 & 2 & 28/50 (56) & $11.22\pm5.21$ & 7.63 & 32.72 \\
 & 5 & 18/50 (36) & $11.11\pm5.37$ & 6.81 & 40.01 \\
 & 9 & 21/50 (42) & $10.37\pm4.74$ & 6.33 & 32.92 \\
\addlinespace[2pt]
\multirow{3}{*}{\ikpd{}}
 & 0 & 11/50 (22) & $12.43\pm5.29$ & 18.86 & 54.37 \\
 & 2 &  9/50 (18) & $12.90\pm3.32$ & 17.43 & 73.94 \\
 & 5 &  0/50 (0)  & ---            &  ---&  ---\\
\midrule
\multicolumn{6}{@{}l}{\textit{0.1\,mm tolerance}} \\
\multirow{5}{*}{Reflex}
 & 0  & 50/50 (100) & $1.71\pm1.02$ & 4.80 & 20.66 \\
 & 2  & 50/50 (100) & $4.02\pm2.97$ & 5.92 & 30.52 \\
 & 5  & 48/50 (96)  & $4.38\pm4.16$ & 5.88 & 25.62 \\
 & 9  & 42/50 (84)  & $5.31\pm4.81$ & 6.45 & 26.86 \\
 & 13 & 29/50 (58)  & $8.64\pm5.78$ & 6.59 & 29.87 \\
\addlinespace[2pt]
\multirow{4}{*}{\ikpd{}}
 & 0 & 48/50 (96) & $6.81\pm4.20$ &  8.89 & 27.74 \\
 & 2 & 46/50 (92) & $7.43\pm3.80$ &  9.88 & 31.80 \\
 & 5 & 38/50 (76) & $7.18\pm4.03$ &  8.18 & 35.84 \\
 & 9 & 26/50 (52) & $8.06\pm4.31$ & 11.16 & 48.23 \\
\bottomrule
\end{tabular*}
\par\smallskip\raggedright\footnotesize
Success: successful/total trials (\%), over the most recent 50 trials per condition. Time is over successful trials (---: none). Force is the norm of the EE force estimated from joint-torque residuals, after removing spikes lasting one or two timestamps. Mean is the within-trial time average, averaged over trials; Peak is the largest filtered value in any trial.
\end{table}

\begin{figure}
    \centering
    \includegraphics[width=\columnwidth]{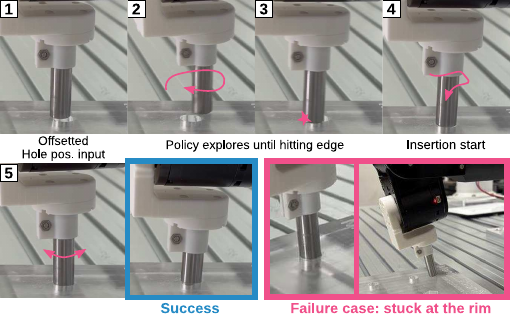}
\caption{Real-world peg insertion with the reflex. (1--5)~Stages of a successful trial. Bottom right: a failed trial, with the peg jammed at the hole entry.}
    \label{fig:insert_real}
    \vspace{-5mm}
\end{figure}

\subsubsection{Experimental setup}

The task is to insert a 12\,mm cylindrical peg, rigidly mounted on the \ac{EE} of the \ac{QDD} arm (Sec.~\ref{sec:robots}, 
Fig.~\ref{fig:opening}B), into holes of the ManipulationNet kit~\cite{chen2026manipulationnet}. Each trial starts with the peg tip 0.5--1\,mm above the plate and offset horizontally from the hole center in a random direction. This initial position is also the hole-position estimate given to the policy, so the offset is the policy's hole-position error and stays fixed during the trial.

The insertion policy follows \ac{FORGE}~\cite{noseworthy2025forge} with two changes: it commands the reflex instead of a Cartesian impedance controller, and the policy does not receive contact-force observations.
An LSTM actor observes the \ac{EE} pose relative to the estimated hole position, the \ac{EE} velocity, a force threshold (10\,N in all experiments), and its previous action. 
At 15\,Hz, it outputs a 3D \ac{EE} position target and a compliance-mode command $m_k\in[0,1]$, with all actions parameterized by a Beta distribution~\cite{petrazzini2021proximal}. The command is thresholded before it reaches the reflex, $m_t=1$ if $m_k>0.5$ and $m_t=0$ otherwise. The commanded orientation is held vertical. The reflex tracks these commands at 50\,Hz, holding the latest command between updates (Fig.~\ref{fig:opening}C).



Training uses holes with 1 and 3\,mm clearance. The policy receives an estimate of the hole center with a random bias of up to 5\,mm, held fixed during each trial. We evaluate at the seen 1.0\,mm clearance and at two unseen clearances, 0.1 and 0.02\,mm, and vary the hole-position error.

We compare three combinations of insertion policy and execution layer: the reflex, \ikpd{}, and Impedance, the last with the gains of \ac{FORGE}.
A separate insertion policy is trained for each, so the methods do not execute identical task-space trajectories and the comparison is between complete systems. For Q2, we also train reflex variants without the spring, plane, or rail primitive, each with its own insertion policy. Simulation results are reported as mean $\pm$ SD over five training seeds of the insertion policy. On hardware, we compare the reflex with \ikpd{}.




\subsubsection{Real-world experiments}
Fig.~\ref{fig:insert_real} shows a real-world trial with the reflex. The peg starts above the erroneous hole estimate and descends onto the plate (1). The policy then searches over the surface in a pattern resembling a classical spiral search~\cite{chhatpar2001search}, learned from training with hole-position error (2). When the peg tip catches the hole edge (3), the policy pushes the peg into the hole (4), wiggling it laterally until it reaches the target depth (5).

We compare the reflex with \ikpd{} on the \ac{QDD} arm at tolerances of 0.1 and 0.02\,mm over hole-position offsets of 0--13\,mm. We report the 50 trials per condition (Table~\ref{tab:insertion_real_world}). The \ac{EE} contact force is estimated from motor currents through joint-torque residuals, identically for both methods.

At 0.1\,mm tolerance, the reflex succeeded in every trial at 0 and 2\,mm offsets and in 84\% at 9\,mm, where \ikpd{} had dropped from 96\% at zero offset to 52\%. The reflex also inserted faster (1.71\,s vs.\ 6.81\,s at zero offset) and with lower estimated force (mean 4.8--6.6\,N vs.\ 8.2--11.2\,N; peak at most 30.5\,N vs.\ up to 48.2\,N).

At 0.02\,mm tolerance, the reflex succeeded in 36--58\% of trials across offsets of 0--9\,mm, whereas \ikpd{} reached at most 22\% and failed in all trials at 5\,mm. Its mean estimated force was less than half that of \ikpd{} (6.3--7.6\,N vs.\ 17.4--20.0\,N), with peaks of at most 40.0\,N vs.\ up to 73.9\,N. In every matched condition, the reflex had a higher success rate and lower mean and peak estimated force.

Across both controllers, 285 of the 294 failed trials (97\%) had entered the hole, i.e., reached at least 2.5\,mm depth, and then jammed during insertion (Fig.~\ref{fig:insert_real}, failure case); only 9 failed to engage the hole.

\subsubsection{Comparative analysis in simulation}
\begin{table}[t]
\caption{Insertion success rates using different controllers.}
\label{tab:downstream_results}
\label{tab:insertion_success_rates}
\centering\small
\setlength{\tabcolsep}{3pt}
\begin{tabular*}{\columnwidth}{@{\extracolsep{\fill}}lrrr@{}}
\toprule
 & \multicolumn{3}{c}{Tolerance (mm)} \\
\cmidrule(l){2-4}
Controller & 0.02$^{\dagger}$ & 0.1$^{\dagger}$ & 1.0 \\
\midrule
Reflex (Ours) & $\mathbf{74.8\pm4.6}$ & $\mathbf{93.1\pm2.8}$ & $\mathbf{99.0\pm0.3}$ \\
IK + PD  & $54.6\pm12.4$ & $72.8\pm7.1$ & $97.9\pm1.6$ \\
Impedance & $10.4\pm4.3$ & $39.7\pm9.9$ & $90.0\pm2.2$ \\
\addlinespace
Ours w/o spring & $68.2\pm1.8$ & $65.0\pm22.7$ & $96.2\pm0.6$ \\
Ours w/o rail & $56.9\pm2.3$ & $76.1\pm6.8$ & $97.8\pm0.2$ \\
Ours w/o plane & $45.9\pm33.7$ & $67.2\pm19.8$ & $96.5\pm3.0$ \\

\bottomrule
\end{tabular*}
\par\smallskip\raggedright\footnotesize
Mean $\pm$ SD over five seeds, 1000 trials per condition, with hole-position offsets of 0--5\,mm. $\dagger$: tolerance not seen in training.
\end{table}

Table~\ref{tab:downstream_results} reports simulated success rates over five seeds at 1.0\,mm, a tolerance seen in training, and at two tighter, unseen tolerances. At 1.0\,mm, the three execution layers were close (99.0\% for the reflex, 97.9\% for \ikpd{}, and 90.0\% for Impedance). The gap widened as the tolerance tightened: the reflex reached $93.1\pm2.8$\% at 0.1\,mm and $74.8\pm4.6$\% at 0.02\,mm, against $72.8\pm7.1$\% and $54.6\pm12.4$\% for \ikpd{}. 

Impedance was the lowest at every tolerance in our experiments. This is likely due in part to its high stiffness gains: we kept the stiffness gains of \ac{FORGE}, so the result should not be read as a limit of impedance control. We examine the effect of stiffness separately in Sec.~\ref{sec:surface_following}.

Removing any one primitive left the 1.0\,mm result nearly unchanged  but lowered success at both unseen tolerances, to 46--68\% at 0.02\,mm and 65--76\% at 0.1\,mm. The ordering of the three variants differs between the two tolerances, and their seed-to-seed variance is large (up to $\pm33.7$\%), so we do not rank the primitives by importance.


\subsubsection{Failure analysis at the hole rim}
To examine where insertion fails, we analyze contact at the hole rim, i.e., above the bore entry, before the peg has entered the hole.
Table~\ref{tab:insertion_rim} reports a separate evaluation with hole-position offsets of 0--3\,mm (four seeds, 2000 trials per method and tolerance). We measure the rim-contact time, averaged over all trials, and count a failure as a \emph{rim failure} if the peg spent most of the final 1\,s at the rim.

The reflex shortened rim contact at both unseen tolerances. At 0.02\,mm, rim failures dropped from 323 to 100 of 2000 trials. This more than accounts for the gain in success, since other failures were more frequent with the reflex (351 vs.\ 236): most \ikpd{} failures occurred at the rim (58\%).
At 0.1\,mm, rim and other failures fell by similar factors (24 vs.\ 55), so the benefit there is not specific to the rim.


\begin{table}[t]
\caption{Rim contact and rim failures in simulated peg insertion, with hole-position offsets of 0--3\,mm.}
\label{tab:insertion_rim}
\centering\small
\setlength{\tabcolsep}{2.9pt}
\begin{tabular*}{\columnwidth}{@{\extracolsep{\fill}}lrrrr@{}}
\toprule
& \multicolumn{2}{c}{Rim-contact time (s)} & \multicolumn{2}{c@{}}{Rim / all failures (\%)} \\
\cmidrule(lr){2-3}\cmidrule(l){4-5}
Tol. (mm) & Reflex & \ikpd{} & Reflex & \ikpd{} \\
\midrule
0.02 & 2.64 & 4.14 & 100/451 (22.2) & 323/559 (57.8) \\
0.1  & 1.59 & 2.88 & 24/123 (19.5)  & 55/306 (18.0) \\
\bottomrule
\end{tabular*}
\par\smallskip\raggedright\footnotesize
Both tolerances are unseen in training. Four seeds, 2000 trials per method and tolerance. Rim-contact time: duration of contact above the bore entry, averaged over all trials. A rim failure is a failed trial that spent most of its final 1\,s at the rim.
\end{table}


\subsection{Surface Following}
\label{sec:surface_following}
\begin{figure}
    \centering
    \includegraphics[width=\columnwidth]{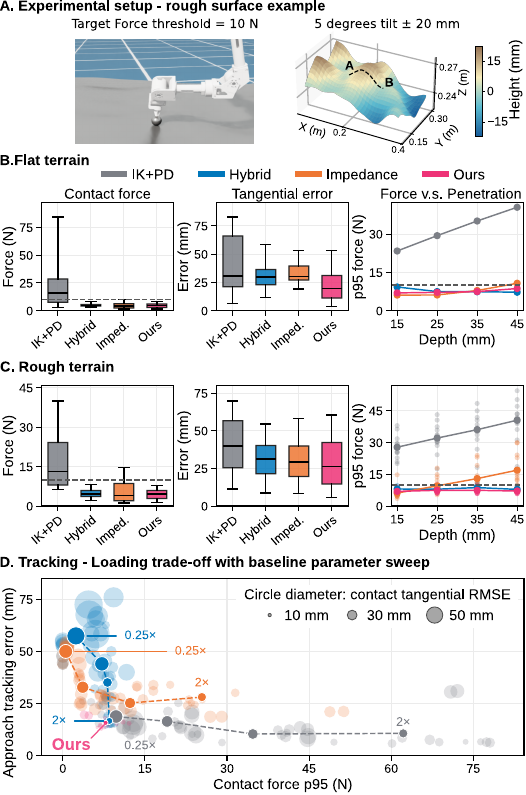}
\caption{Simulated surface following results. (A)~Setup and an example rough surface. (B,~C)~Flat and rough surfaces: contact force, tangential error, and force p95 versus commanded penetration. Boxes show the median, interquartile range, and 5th--95th percentiles during contact. Dashed lines mark $F_{\mathrm{th}}=10$\,N. (D)~Force p95 versus free-space position error; circle size indicates tangential RMSE. Baselines span stiffness scales of $0.25\times$ to $2\times$; the reflex is fixed.}
    \label{fig:wiping_results}
    \vspace{-5mm}
\end{figure}

Surface-following tasks such as wiping and deburring require the tool to slide along a surface while pressing on it with a bounded force.
In practice, the surface geometry is known only approximately, so the commanded path lies partly below the surface. 
We reproduce this situation by commanding a trajectory below the nominal surface, over terrain whose height is unknown to the controller (Fig.~\ref{fig:wiping_results}A).

Now there is conflict between tracking the command and limiting the contact force, while motion along the surface should be unaffected. Stiff position control resolves this conflict in favor of tracking and soft control in favor of low loading, which makes the task a direct test of the tradeoff that gain tuning moves along. The study runs in simulation with the \ac{QDD} arm and addresses Q1, with commanded penetration and unknown surface height as the command error, and Q2, through a gain sweep of the classical baselines.



\subsubsection{Experimental setup}
We test on one flat surface and 12 rough surfaces generated with Perlin noise, with height variations of $\pm20$\,mm that are unknown to the controllers (Fig.~\ref{fig:wiping_results}A). All controllers receive the same nominal trajectory: a 2\,s approach followed by two 8\,s horizontal cycles, at a commanded penetration of 15, 25, 35, or 45\,mm below the nominal surface.

We compare the reflex with \ikpd{}, Impedance, and Hybrid. The reflex was trained with $F_{\mathrm{th}}=10$\,N and is not tuned for this task. Its compliance mode is switched by a simple contact rule: $m_t=1$ while the estimated contact force exceeds 1\,N, and $m_t=0$ otherwise.
Hybrid regulates the force along the fixed world $z$-axis to 5\,N, using a force estimate from joint torques, and controls motion in the remaining directions. On rough surfaces, this axis need not coincide with the local surface normal.

We report the 95th percentile (p95) of the contact force and the tangential tracking RMSE, both during surface following, and the position RMS error during the free-space approach.

\subsubsection{Contact loading and surface adaptation}

The reflex matched Hybrid in contact loading and tangential tracking on both surface types (Fig.~\ref{fig:wiping_results}B,~C). On the flat surface, the contact-force p95 was 7.5\,N for the reflex and 7.4\,N for Hybrid. On rough surfaces, it was 7.4 and 8.7\,N, with tangential RMSE of 31.3 and 32.6\,mm, respectively.

As commanded penetration increased from 15 to 45\,mm, the loading of the reflex and Hybrid changed little and stayed below the 10\,N threshold. In contrast, the loading of \ikpd{} rose with penetration on both surface types, and Impedance became sensitive to penetration on rough surfaces. \ikpd{} also had the largest tangential errors, although it is the stiffest of the controllers: stiffer position tracking does not improve motion along a constrained surface.


The reflex thus absorbs the mismatch between the commanded trajectory and the real surface along the normal direction, at little in tangential motion relative to Hybrid. Unlike Hybrid, it is given neither a force-control axis nor a force setpoint.


\subsubsection{Tracking--loading balance and tuning}
Figure~\ref{fig:wiping_results}D compares the fixed reflex with the baselines at stiffness scales $s\in\{0.25,0.5,1,2\}$. For every baseline, the stiffness gains are scaled by $s$ and the damping gains by $\sqrt{s}$, which keeps the damping ratio constant, as in Franka's Cartesian impedance example~\cite{franka_cartesian_impedance_example}.

For \ikpd{} and Impedance, stiffness trades approach accuracy against contact loading: stiffening reduced the approach error but raised the loading, and softening did the reverse. Neither direction improved tangential tracking. No tested setting reached the region occupied by the reflex, and only Hybrid, at its best stiffness scales, came close. 

 This answers the classical side of Q2: gain tuning alone does not reproduce the benefit. Hybrid sets the reference in this task, because its force-control axis and force setpoint give it the constraint structure in advance. The reflex comes close to this reference without being given either.




\section{Conclusion}

We showed that the low-level control layer beneath a task-level command source strongly affects contact-rich performance, and that a proprioceptive reflex can fill this role. The reflex is trained in simulation on three simple interaction primitives and runs on top of a compliant joint-level controller. Without observing forces or constraint geometry, it limited contact forces and sustained task success under imperfect commands and uncertain geometry. On hardware, it completed dual-arm box lifting at every commanded penetration depth, whereas \ikpd{} diverged at the largest one. In peg insertion at 0.02\,mm tolerance, it raised success from at most 22\% to 36--58\% with less than half the mean estimated contact force. In simulated surface following over unknown terrain, its tradeoff between free-space tracking and contact loading was close to that of hybrid motion--force control, although it was given neither a force-control axis nor a force setpoint. Gain tuning of \ikpd{} and Impedance did not reach this tradeoff, and removing any one primitive from training lowered simulated insertion success at the unseen tolerances.

Future work will address three limitations. First, evaluation on a broader range of contact-rich tasks is needed to establish how far the approach generalizes. Second, the interaction primitives model translational loading and constraints only. They do not capture coupling between translation and rotation, such as opening a door with a rigid grasp. We plan to investigate rotational primitives in the future. Third, the reflex depends on an explicit compliance-mode input to distinguish nominal tracking from compliant interaction. This input must come from the high-level command source or from separate mode-selection logic, such as the force-threshold rule in Sec.~\ref{sec:surface_following}, which complicates integration with policies that output only motion commands. We will investigate two ways to remove this requirement: having the high-level agent learn the mode together with its motion commands, as the insertion policy in Sec.~\ref{sec:downstream} already does, and letting the reflex adapt its compliance online from its proprioceptive history.

\section*{Acknowledgments}
The authors used AI tools (ChatGPT, Codex, and Claude) to draft and edit text throughout the manuscript based on the authors' own content, data, and results.
The research ideas, method, experimental design, experiments, data, and figures are the authors' own; plots were generated by the authors from the study's data using standard libraries such as Matplotlib. The tools generated no experimental data or results. All references were checked by the authors against the original sources. The authors reviewed all AI-assisted content and take full responsibility for the manuscript.

\bibliographystyle{IEEEtran}
\bibliography{references}
\end{document}